\documentclass{article}

 \usepackage[nonatbib,preprint]{neurips_2023}

\usepackage[utf8]{inputenc} 
\usepackage[T1]{fontenc}    
\usepackage{hyperref}       
\usepackage{url}            
\usepackage{booktabs}       
\usepackage{amsfonts}       
\usepackage{nicefrac}       
\usepackage{microtype}      
\usepackage{xcolor}         

\usepackage{enumerate}
\usepackage{multirow}
\usepackage[export]{adjustbox}
\usepackage{authblk}

\title{BrainCLIP: Bridging Brain and Visual-Linguistic Representation Via CLIP for Generic Natural Visual Stimulus Decoding}

 \author[ ]{Yulong Liu${^1}$}
 \author[ ]{Yongqiang Ma${^1}$}
 \author[ ]{Wei Zhou${^1}$}
\author[ ]{Guibo Zhu${^2}$}
\author[ ]{Nanning Zheng${^1}$\thanks{Corresponding author}}

\affil[1]{Institute of Artificial Intelligence and Robotics, Xi'an Jiaotong University,  Xi'an, China }
\affil[2]{Institute of Automation, Chinese Academy of Sciences (CASIA), Beijing,  China}
\affil[ ]{\textit{\{lylhubxy, zw85231507\}@stu.xjtu.edu.cn}}
\affil[ ]{\textit{musayq@xjtu.edu.cn}}
\affil[ ]{\textit{gbzhu@nlpr.ia.ac.cn}}
\affil[ ]{\textit{nnzheng@mail.xjtu.edu.cn}}

\begin{document}

\maketitle

\begin{abstract}
    Due to the lack of paired samples and the low signal-to-noise ratio of functional MRI (fMRI) signals, reconstructing perceived natural images or decoding their semantic contents from fMRI data are challenging tasks. In this work, we propose, for the first time, a task-agnostic fMRI-based brain decoding model, BrainCLIP, which leverages CLIP's cross-modal generalization ability to bridge the modality gap between brain activity, image, and text. Our experiments demonstrate that CLIP can act as a pivot for generic brain decoding tasks, including zero-shot visual categories decoding, fMRI-image/text matching, and fMRI-to-image generation. Specifically, BrainCLIP aims to train a mapping network that transforms fMRI patterns into a well-aligned CLIP embedding space by combining visual and textual supervision. Our experiments show that this combination can boost the decoding model's performance on certain tasks like fMRI-text matching and fMRI-to-image generation. On the zero-shot visual category decoding task, BrainCLIP achieves significantly better performance than BraVL, a recently proposed multi-modal method specifically designed for this task. BrainCLIP can also reconstruct visual stimuli with high semantic fidelity and establishes a new state-of-the-art for fMRI-based natural image reconstruction in terms of high-level semantic features.
  
\end{abstract}

\bibliographystyle{unsrt}
\section{Introduction}
Brain decoding enables the interpretation of mental content,
which is a crucial technology for brain-computer interface and
has long been a hot topic in the field of neuroscience. To
achieve the goal of reading the human mind, functional magnetic
resonance imaging (fMRI) is widely used as a non-invasive
approach to decipher the perception and semantic information encoded in the cerebral cortex. Among the diverse fMRI-based brain reading tasks, decoding natural visual information is one of the most important but challenging problems since visual information is most relevant for daily perception while rich in statistical structure and semantic content\cite{field1987relations,karklin2009emergence}. 

In the early stage of visual stimulus decoding, research largely relied on a classification-based approach where a multi-voxel fMRI pattern of brain activity was used to determine the class of the stimulus drawn from a predefined set of categories\cite{haxby2001distributed,carlson2003patterns,cox2003functional,haynes2005predicting,kamitani2005decoding}.  Subsequently, Horikawa and Kamitani \cite{horikawa2017generic} proposed generic object decoding where training and testing categories had no overlap, and a group of linear regression models was trained to characterize the relationship between brain activity and visual features extracted by a pre-trained CNN. Recently, Du et al. \cite{du2023decoding} proposed a method based on multi-modality joint learning to address the zero-shot neural decoding problem, but there remains much room for improvement.

A relatively harder problem is natural visual stimulus reconstruction which requires the model to reveal the structure and semantic content of the images simultaneously. Earlier works tend to emphasize pixel-level similarity with the original images\cite{beliy2019voxels,shen2019end,shen2019deep,lin2019dcnn,ren2021reconstructing}, which usually results in blurry and unintelligible reconstructions, while more and more recent studies focus on producing recognizable or semantically meaningful scenes\cite{mozafari2020reconstructing,ozcelik2022reconstruction,lin2022mind,takagi2022high,ozcelik2023brain}. Most of the recent advances in the field of visual stimulus reconstruction are attributed to the development of deep generative models like GAN and diffusion model.  However, existing methods often generate reconstructions that are unfaithful to the actual stimuli. The perfect reconstruction for natural visual stimuli is still an open problem.
    
    Although many methods and elaborate model architectures have been proposed for fMRI-based brain decoding, there is no one can generalize well to all tasks.  In this paper, we return to the essence of brain decoding, i.e.,  representation learning, which is crucial for successful brain decoding. We propose BrainCLIP, a new brain decoding framework,  which leverages CLIP's cross-modal generalization ability to directly connect brain activity with image and text by combining visual and textual supervision. Our framework is inspired by the successful application of CLIP\cite{radford2021learning} in various domains \cite{mokady2021clipcap,kim2022diffusionclip,tewel2022zerocap}.
    These previous studies suggest that the visual encoders of CLIP, which were pre-trained with 400M image-text pairs, can capture both the semantic and structural information of an image. Further, according to Lu et al. \cite{lu2022multimodal}, both visual and textual encoders trained multimodally are more brain-like than unimodal ones. Thus, we leverage the well-aligned semantic space of CLIP to perform neural decoding tasks.
    \begin{figure*}
  \centering
\centerline{\includegraphics[width=1.\textwidth]{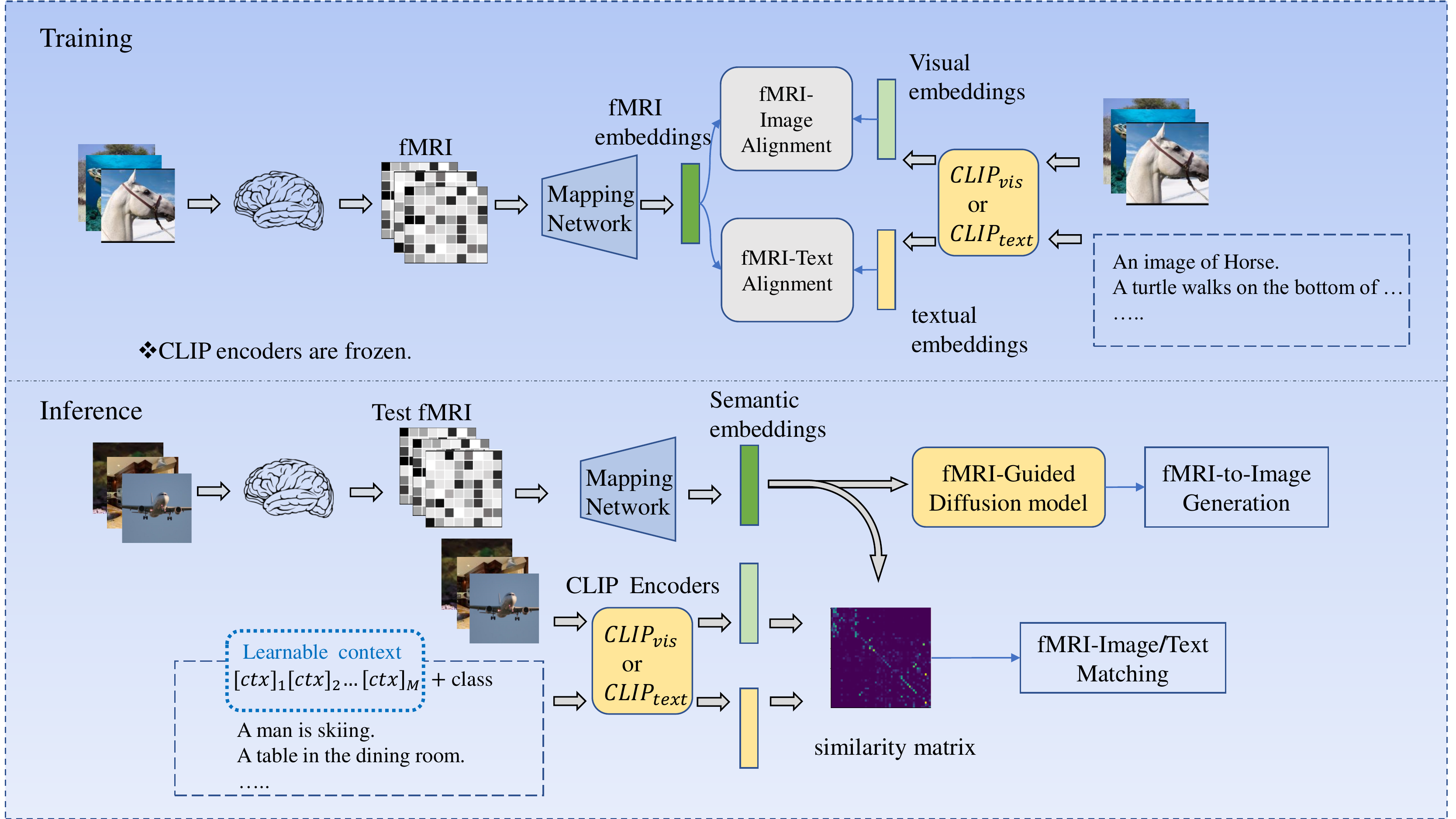}}
\caption{An overview of our BrainCLIP framework. In the training phase, a mapping network is trained to align fMRI patterns with images and texts. In the test phase, BrainCLIP can be flexibly applied to the fMRI-image/text matching or fMRI-to-image generation tasks. Combined with prompting techniques, BrainCLIP can also conduct zero-shot visual categories classification.   }
\label{overview}
\end{figure*}

    The overview of our framework is shown in Figure \ref{overview}. In the training phase, an fMRI mapping network is trained to align fMRI data with visual and textual data via contrastive learning.
    In the inference phase, we can flexibly apply BrainCLIP to the fMRI-Image/Text-matching task or fMRI-to-image generation task. We also successfully apply BrianCLIP to zero-shot visual categories classification by combining it with prompting techniques.
    Our main contributions are summarized as follows:
\begin{enumerate}[(1)]
    \item We propose BrainCLIP, the first task-agnostic neural decoding model, which makes it possible to directly connect brain activities with images and texts and can be flexibly applied to related tasks. It obtains good zero-shot neural classification accuracy and establishes a new state-of-the-art for fMRI-based natural image reconstruction in terms of high-level image features. It is possible to combine BrainCLIP with future brain decoding models by using it as a global similarity constraint. This work is also the first one to report large-scale fMRI-image/text retrieval and shows that it is feasible to decode visual-linguistic representation from the visual cortex with high accuracy.
    \item  Our experiments demonstrate that the combination of fMRI-Image alignment and fMRI-text alignment can lead to better semantic decoding performances in comparison with the conditions where either modality exists alone.   
    \item Two different architectures (a linear version and a VAE-based version) for the BrainCLIP's mapping network are explored and extensive experiments are conducted to evaluate their performances.    
\end{enumerate}
\section{Related works}
\paragraph{Zero-shot neural decoding of visual categories} Earlier classification-based neural decoding methods \cite{haxby2001distributed,carlson2003patterns,cox2003functional,haynes2005predicting,kamitani2005decoding} usually used a classifier to learn the relationship between brain activity and a predefined set of labels. The resulting models cannot be applied to new classes. Then, Horikawa and Kamitani \cite{horikawa2017generic} proposed to decode arbitrary object categories by learning a linear regression function that transformed fMRI patterns into CNN features of the viewed images. The novel image categories can be estimated by comparing the correlation coefficients between the predicted image features and the category-average visual features of various categories. However, the estimation performance of this method significantly depends on a large number of paired stimuli-responses data. And the multi-modal semantic knowledge underlying brain activities is under-explored. Akamatsu et al.\cite{akamatsu2020brain} proposed a semi-supervised multi-view Bayesian generative model that mapped brain activity, visual features, and category features (word2vec\cite{mikolov2013distributed} representations) into a shared latent representation. In the test phase, the visual features and category features were reconstructed from the latent representations predicted from the test fMRI patterns. However, the linear hypothesis in the proposed model does not accord with the actual visual encoding-decoding process in the human brain. Recently, Du et al. \cite{du2023decoding}  proposed BraVL, a VAE-based model that works in two collaborative parts, i.e., multi-modality joint modeling and mutual information regularization, to learn the relationship between brain activities, visual stimuli, and Wikipedia descriptions of classes. Then, a support vector machine is used to perform the classification task with the learned latent representation. Our method is most related to BraVL but not restricted to the classification task. And the textual data used in our experiments are image captions instead of class descriptions.
\paragraph{Visual stimulus reconstruction} A number of methods have been proposed for reconstructing a visual stimulus from fMRI. From the aspects of reconstruction results, existing methods can be categorized into two groups: one group tends to emphasize pixel-level similarity with the original images\cite{beliy2019voxels,shen2019end,shen2019deep,lin2019dcnn,ren2021reconstructing,9892276}, the other group focuses on producing recognizable or semantically meaningful scenes\cite{mozafari2020reconstructing,ozcelik2022reconstruction}.  From the aspects of the used methods, there also exist two main streams: regression-based methods\cite{naselaris2009bayesian,nishimoto2011reconstructing,shen2019deep,mozafari2020reconstructing,ozcelik2022reconstruction,lin2022mind,takagi2022high,ozcelik2023brain} and End-to-end Deep Learning methods\cite{st2018generative,seeliger2018generative,shen2019end,lin2019dcnn,beliy2019voxels,ren2021reconstructing}.
Regression-based methods usually use a linear regression model to convert fMRI patterns into intermediate representations (handcrafted or DNN-based image features) and then use a decoder to reconstruct the visual stimuli from the intermediate representations. End-to-end Deep Learning methods attempt to directly decode an fMRI recording into its corresponding image stimulus using high-complexity deep models.
\paragraph{Cross-modal contrastive learning}Cross-modal contrastive learning aims to align data from different modalities (especially, from visual and textual modalities) through contrastive loss (e.g., InfoNCE loss\cite{oord2018representation}). In recent years, such an alignment strategy has obtained great success on vision-and-language downstream tasks\cite{radford2021learning,jia2021scaling,NEURIPS2022_a90b9a09,NEURIPS2022_6a386d70}. Furthermore, a recent study\cite{lu2022multimodal} reported that
 the visual or lingual features extracted by a multi-modal model pre-trained with contrastive loss are more predictive of the corresponding brain activities than the features extracted by a unimodal model, which indicates that cross-modal contrastive learning is a more brain-like learning method than those unimodal methods. 
 Several neuroscience studies also revealed that concrete concepts are encoded in the brain both visually and linguistically\cite{paivio2013imagery,bi2021dual,jing2022exploring}.
 Inspired by these previous works, we use CLIP\cite{radford2021learning}, the pioneering work of vision-language contrastive learning, as a bridge to connect brain activities with images and textual descriptions. And cross-modal contrastive learning is also adopted in our training process.
 \paragraph{Diffusion probabilistic model}
Diffusion probabilistic models consist of a forward diffusion process
and a reverse diffusion process. The forward process is a Markov chain that follows a Gaussian
transition $ q(x_t|x_{t-1}):=\mathcal{N}(\sqrt{1-\beta_t}x_{t-1},\beta_t\bold{I})$, where ${\{\beta_t\}}_{t=0}^T$ are fixed or learned variance schedule. In the  forward process, noise is gradually added to the data when sampling the latent variables $x_t$ for $t=1,\dots, T$.  The resulting
latent variable $x_t$ can be expressed as:
\begin{equation}\label{eq0}
    x_t=\sqrt{\alpha_t}x_{0}+\sqrt{1-\alpha_t}\mathcal{\epsilon},  \epsilon\sim\mathcal{N}(\bold{0,I})
\end{equation}
,where $\alpha_t:=\prod_{s=1}^t(1-\beta_s)$.  The reverse process is  also parametrized by a Gaussian transition $p_\theta(x_{t-1}|x_t):=\mathcal{N}(x_{t-1};\mu_\theta(x_t,t),\sigma_\theta(x_t,t)\bold{I})$. On this basis, Ho et al.\cite{ho2020denoising} proposed denoising  diffusion probabilistic models (DDPM) that decompose $\mu_\theta(x_t,t)$ into the linear combination of $x_t$ and a noise
approximation model $\epsilon_\theta(x_t, t)$, which can be learned by
solving the following optimization problem:
\begin{equation}\label{eq1}
\mathop{min}\limits_{\theta} \mathbb{E}_{x_0 \sim q(x_0),\epsilon \sim \mathcal{N}(\bold{0,I}),t}{|| \epsilon-\epsilon_\theta(x_t,t)||_2^2}.
\end{equation}
 After training $\epsilon_\theta(x,t)$, the samples are generated from following reverse diffusion process:
\begin{equation}\label{eq2}
x_{t-1}=\frac{1}{\sqrt{1-\beta_t}}(x_t-\frac{\beta_t}{\sqrt{1-\alpha_t}}\epsilon_\theta(x_t,t))+\sigma_{t}z,  z\sim\mathcal{N}(\bold{0,I})
\end{equation}
 It was found that the sampling process
of DDPM has a relationship with that of the score-based generative models\cite{song2019generative,song2020score}, which is given as follows:
\begin{equation}\label{eq3}
  \epsilon_\theta(x_t,t)=-\sqrt{1-\alpha_t}\bigtriangledown_{x_t}logp_\theta(x_t).
\end{equation}


\section{Materials and Methods}
  \subsection{Experimental Data}
  \paragraph{Generic Object Decoding(GOD) Dataset}
   GOD dataset was curated by Horikawa and Kamitani\cite{horikawa2017generic}, which consists of the brain fMRI recordings of five healthy subjects presented with images from ImageNet \cite{deng2009imagenet}. The GOD comprises 1250 distinct images drawn from 200 ImageNet categories, among which 1200 training images are from 150 categories (150x8) and 50 test images from the remaining 50 categories.   The training and test image stimuli were presented to the subjects 1 and 35 times, respectively, leading to 1200 and 1750 fMRI instances. In this work, we use the preprocessed regions of interest (ROI)\footnote{The preprocessed data and demo code are available at \url{http://brainliner.jp/data/brainliner/Generic_Object_Decoding}}, which covered the voxels from the early visual cortex to higher visual areas. For each test image, the fMRI responses from different trials are averaged.  In addition to fMRI and image data, we manually annotated a caption for each image in GOD, which contains the foreground and background information. And we also used a template to embed the class names into sentences. Therefore, every visual stimulus has two textual descriptions, one of which is randomly chosen as training data in every epoch.  
  \paragraph{Natural Scenes Dataset (NSD)} Unlike GOD where the stimuli are object-centered images, NSD\cite{allen2022massive} contains the images of more complexe natural scenes. NSD  was collected from 8 subjects viewing images from the COCO dataset\cite{lin2014microsoft}. In this work, we conducted experiments using the data from subjects 1, 2, 5, and 7 in NSD. The training set for each subject contains 8859 image stimuli and 24980 fMRI trials (up to 3 trials for each image). The test set contains 982 image stimuli and 2770 fMRI trials.  For the images that have multiple fMRI trials, their responses are averaged across different trials.
  Using the provided NSDGeneral ROI mask in 1.8 mm resolution, we get the ROIs for the 4 subjects, which consist of 15724, 14278, 13039, and 12682 voxels respectively, and cover visual areas from the early visual cortex to higher visual areas. 
  Corresponding captions can be extracted from the COCO dataset according to the ID of the image stimuli. Multiple captions may correspond to the same image, and one of them is randomly chosen as training data in each epoch. 
  
\subsection{Training}
 \paragraph{Cross-Modal Alignment}
  The essential idea of cross-modal alignment is to pull the embedding vectors of paired samples close while pushing those of unpaired samples far apart. It's an approach to maximize the mutual information between two matched samples, which are assumed to describe the same semantic meaning but from different modalities. Since the number of fMRI-image pairs is usually limited, which may be insufficient for meaningful representation learning, we propose to combine fMRI-image alignment and fMRI-text alignment. The alignment is achieved by minimizing contrastive  loss. Given a batch of embeddings $A=\{A_i\}_{i=0}^M$ for data from modality$\mathcal{A}$, and $B=\{B_i\}_{i=0}^M$ for data from modality $\mathcal{B}$, where $M$ is the batch size, for each $A_i$,  $B_i$ is defined as its positive sample and the others in $\{B_j\}_{j\neq i}$ are defined as negative samples.  We use the infoNCE definition\cite{oord2018representation} of contrastive loss:
\begin{equation}\label{eq4}
Contrast(A,B;\tau)=-\frac{1}{M}\sum_{i=1}^{M}log[\frac{exp({cos(A_i,B_i)}/{\tau})}{\sum_{j=1}^M exp( {  cos(A_i,B_j)}/{\tau}          )}], 
\end{equation}
where $\tau$ is a temperature hyper-parameter.
During training, given a batch of data, the fMRI mapping network encodes fMRI patterns to fMRI embeddings $ F=\{F_i\}_{i=0}^M $, the CLIP visual encoder encodes images to image embeddings $I=\{I_i\}_{i=0}^M$, and the CLIP textual encoder encodes texts to textual embeddings $T=\{T_i\}_{i=0}^M$. We define
\begin{equation}\label{eq5}
  L_{FI}=\frac{1}{2}[Contrast(F,I;\tau_1)+Contrast(I,F;\tau_1)], 
\end{equation}
\begin{equation}\label{eq6}
L_{FT}=\frac{1}{2}[Contrast(F,T;\tau_2)+Contrast(T,F;\tau_2)]. 
\end{equation}

 The training objective is:
 \begin{equation}\label{eq7}
  L=\alpha( L_{FI})+(1-\alpha) L_{FT},
\end{equation}
where $\alpha$, $\tau_1$ and $\tau_2$ are non-negative hyper-parameters selected through sweeps. 
 \paragraph{Mapping Network} To align fMRI data with corresponding images and texts, two kinds of mapping networks are explored in this work, one is a simple full connection layer (linear layer) and the other is a variational autoencoder(VAE)\cite{shi2019variational}, which consists of an fMRI encoder and a decoder. Specifically, before training the VAE-based fMRI mapping network, we jointly pre-trained the VAE decoder and a visual encoder with large-scale unpaired images( the natural images from the validation set of ILSVRC2012\cite{deng2009imagenet} and the training set of MSCOCO2014\cite{lin2014microsoft}). The visual encoder can encode a CLIP visual embedding vector into a latent representation while the decoder can reconstruct the visual embedding from the latent representation. Then, the VAE decoder is frozen and only the fMRI encoder needs to be trained.  We refer to the two variants of BrainCLIP as \textbf{BrainCLIP-Linear} and \textbf{BrainCLIP-VAE} respectively.  During training,  weight decay is applied on both the two kinds of architectures, and the CLIP encoders are frozen. More implementation details can be found in the Appendix.

\subsection{Inference}
 In the inference phase, the fMRI  mapping network is used to transform fMRI signals into semantic embedding vectors. Combined with CLIP's visual encoder or textual encoder, BrainCLIP can be flexibly applied to the fMRI-image/text matching task and fMRI-to-image generation task.
 \paragraph{fMRI-to-text retrieval \& zero-shot classification via prompting}
 BrainCLIP can directly measure the semantic distance between an fMRI pattern and a text description by calculating the cosine similarity of their embeddings, which makes it easy to be applied to the text retrieval task, the goal of which is to find the correct text description from a candidate pool. Furthermore, large pre-trained vision-language models like CLIP have recently shown great potential in learning generic visual representations and allowing zero-shot transfer to various downstream classification tasks via prompting\cite{radford2021learning,jia2021scaling,NEURIPS2022_6a386d70,zhou2022learning}. By leveraging CLIP's well-aligned embedding space, BrainCLIP can also perform zero-shot visual stimulus classification. This is achieved by comparing fMRI embeddings with the classification weights synthesized by CLIP's text encoder, which takes as input textual prompts specifying classes of interest. The textual prompts can be manually designed like  “a photo of a [CLASS].” where the class token is replaced by the specific class name and the other tokens play a role in providing a context for the class name. Further, Zhou et al.\cite{zhou2022learning} proposed to automatically learn the embedding vectors for the context tokens instead of manually designing the prompts, which is referred to as Context Optimization (CoOp). BrainCLIP can work in both two ways. When conducting CoOp, we use the context vectors pre-trained on ImageNet, which is released by Zhou et al\footnote{\url{https://github.com/KaiyangZhou/CoOp}}.
 
 \paragraph{fMRI-to-image retrieval} The goal of the fMRI-to-image retrieval task is to find the correct visual stimulus from a pool of candidate images given an fMRI pattern as the query. BrainCLIP can achieve such a function by comparing the cosine distance between fMRI embedding predicted from an fMRI pattern and those of the candidate images. 
 \begin{figure*}
  \centering
\centerline{\includegraphics[width=0.7\textwidth]{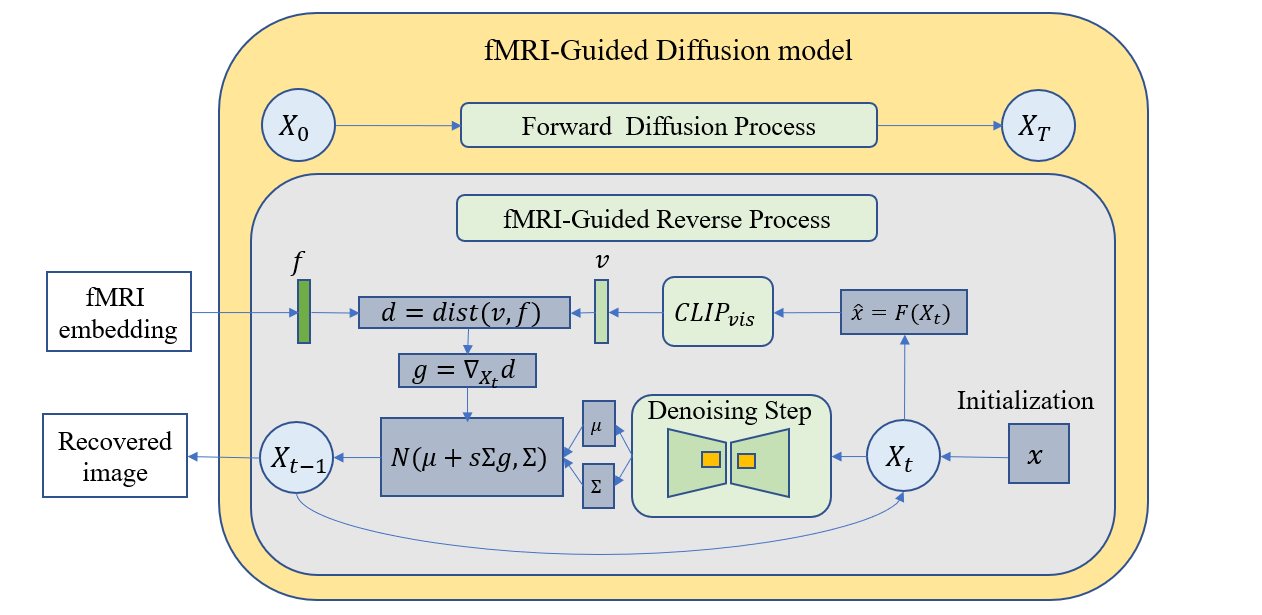}}
\caption{fmri-guided diffusion model}
\label{fmri_guided}
\end{figure*}
\paragraph{fMRI-to-image generation with fMRI-guided diffusion model}
Diffusion model\cite{sohl2015deep,ho2020denoising} is currently a state-of-the-art generative method that can generate realistic images. Several works have explored conditional image synthesis with guided diffusion models\cite{dhariwal2021diffusion,kim2022diffusionclip}. Based on the relationship between DDPM and score-based generative models, Dhariwal and Nichol\cite{dhariwal2021diffusion} proposed a classifier-guided diffusion model, where a classifier $p_\psi(y|x_t, t)$ was trained and its gradients $\bigtriangledown_{x_t}log p_\psi(y|x_t,t)$ was used to guide the diffusion sampling process towards an arbitrary class label $y$. 
Given an input state $x_t$ and current time step $t$, if the predicted mean and variance of $x_{t-1}$ after an unconditional denoising step are $\mu$ and $\Sigma$, then $x_t$ can be sampled from the following distribution:
\begin{equation}\label{eq8}
\mathcal{N}(\mu+s\Sigma\bigtriangledown_{x_t}log p_\psi(y|x_t,t),\Sigma)
\end{equation}
, where $s$ is a positive guidance scale. 
The derivation can be found in \cite{dhariwal2021diffusion}. In our work, we replace the classifier with  BrainCLIP, whose gradients are used to iteratively optimize the generated image intending to minimize the squared spherical distance between the CLIP embedding of the reconstructed image and the fMRI embedding. The pipeline of the fMRI-guided diffusion model is shown in Figure \ref{fmri_guided}, where  $\hat{x}$ is a linear combination of estimated  $x_0$ and current state  $x_t$. 
\begin{equation}\label{eq10}
 \hat{x}=\hat{x}_0\eta+x_t(1-\eta),
\end{equation}
in our experiments, $\eta=\sqrt{1-\alpha_t}$.

By reversing equation (\ref{eq0}), we get:
\begin{equation}\label{eq9}
 \hat{x}_0=\frac{x_t-\sqrt{1-\alpha_t}\epsilon_\theta(x_t,t)}{\sqrt{\alpha_t}}.
\end{equation}
At the beginning of the fMRI-guided reverse process,  $x_t$ can be initialized with either a Gaussian noise or a noised image. For the latter, we prepare a set of natural images as the image prior, which contains 170k images from MSCOCO and ImageNet. For each test fMRI pattern, an image with the closest semantic distance is chosen by BrainCLIP from the image prior and set as the initial state after added noise. It is worth noting that the proposed conditioning process is independent of the training of the diffusion model. And the diffusion model used in this work is pre-trained
on ImageNet\cite{deng2009imagenet} and provided by Dhariwal and Nichol\footnote{ \url{https://github.com/openai/guided-diffusion} }.

\begin{table*}
     \setlength{\abovecaptionskip}{0pt}%
\setlength{\belowcaptionskip}{10pt}%
    \centering
      \caption{\textbf{Large-scale fMRI-to-image retrieval \& fMRI-to-text retrieval on NSD dataset}. The number of candidates is 982. Chance levels for Recall@1, @5,and @10 are 0.10\%,0.51\% and 1.01\% respectively.  V\&T means that the model is trained with visual and textual features while V(T) means that the model is trained with only visual(textual) features. }
    \resizebox{\linewidth}{!}{
    \begin{tabular}{ccccccc}
        \toprule
        Task&Methods& Modality&Recall@1 &Recall@5 &Recall@10 &Mean Recall \\
        \midrule
          \multirow{8}*{fMRI-to-Image Retrieval}&Baseline-Ridge&V&12.78&34.80&48.55&32.04\\
         &Baseline-Ridge &T&8.38&23.57&36.28&22.74\\
          \cmidrule(r){2-7}
        &BrainCLIP-Linear&V\&T&24.62&55.52&68.91&49.68\\
        &BrainCLIP-Linear&V &\textbf{27.47}&\textbf{57.06}&\textbf{71.10}&\textbf{51.88}\\
        &BrainCLIP-Linear&T&10.57&29.89&44.81&28.42\\
       \cmidrule(r){2-7}
        &BrainCLIP-VAE &V\&T&17.49 &42.46 &57.13 &39.03 \\ 
         &BrainCLIP-VAE&V&17.21&42.90&57.20&39.10\\
          &BrainCLIP-VAE &T&11.58&32.43&46.36&30.12\\
        \midrule
        \multirow{8}*{fMRI-to-Text Retrieval}&Baseline-Ridge&V&5.52&18.51&29.76&17.93\\
        &Baseline-Ridge &T&6.44&20.80&32.54&19.93\\
        \cmidrule(r){2-7}
         &BrainCLIP-Linear&V\&T&10.16&28.59&43.20&27.32\\ 
         &BrainCLIP-Linear&V&6.52&20.21&30.58&19.10\\
        &BrainCLIP-Linear&T&8.94&26.02&39.61&24.86\\
      \cmidrule(r){2-7}
        &BrainCLIP-VAE &V\&T&\textbf{10.44} &\textbf{31.06}  &\textbf{45.19} &\textbf{28.90}\\ 
         &BrainCLIP-VAE&V&7.43&24.49&37.60&23.17\\
          &BrainCLIP-VAE&T&9.70&30.65&45.16&28.50\\

      \bottomrule
      \end{tabular}}
       \label{retrieval}
\end{table*}

\section{Results}
\subsection{Large-scale fMRI-to-image retrieval \& fMRI-to-text retrieval}
We report Recall@K (recall of top K candidates) with K = 1, 5, 10 for
both  fMRI-to-image and fMRI-to-text retrieval on the NSD dataset. The average of Recall@K, i.e., Mean Recall, is used for the final comparison. For each test fMRI pattern, the candidate pool consists of the 982 test images or 982 image captions(chance levels for Recall@1, @5, and @10 are 0.10\%, 0.51\%, and 1.01\% respectively). Since it is the first work to report fMRI-image/text matching on such a large scale, there are no previous works to compare with, we consider ridge regression as our baseline. The ridge regression models are trained with either CLIP visual embedding or CLIP textual embedding(using mean squared error as the loss function). All the results are averaged across the 4 subjects.
From the results shown in Table\ref{retrieval}, we can get the following observations. First, BrainCLIP gets significantly better performance than baseline on both fMRI-to-image and fMRI-to-text retrieval tasks. Second, the combination of visual and textual supervision can improve the fMRI-to-text retrieval success rate for both the two variants of BrainCLIP, but it can drop the fMRI-to-image retrieval success rate in comparison with the condition where only visual supervision exists, especially for BrainCLIP-Linear. Third, when only textual supervision exists, BrainCLIP-VAE has better performance than BrainCLIP-Linear on both fMRI-to-image and fMRI-to-text retrieval tasks. But if visual supervision exists, BrainCLIP-Linear has significantly better performance on the fMRI-to-image retrieval task, while BrainCLIP-VAE is better at the fMRI-to-text retrieval task. These results imply that BrainCLIP-VAE can capture more high-level semantic-related information while BrainCLIP-Linear may distinguish images via some lower-level features.
 
\subsection{Zero-shot classification by prompting}
We evaluate BrainCLIP's zero-shot classification performance on the GOD dataset.
In the test set of GOD, there are 50 visual stimuli from 50 categories that have no overlap with the training set, which requires that the models can generalize to unseen classes. Two ways of prompting are compared, i.e., manually designed templates versus CoOp. Since there are three pre-trained CoOp weights with a context token length of four, their scores are averaged as the final results. 
The top-1 and top-5 classification accuracy (with the chance levels of 2\% and 10\%, respectively) are compared with several representative methods in Table \ref{zero_classfication}.  
\begin{table*}
    \setlength{\abovecaptionskip}{0pt}%
\setlength{\belowcaptionskip}{10pt}%
     \caption{\textbf{Zero-shot visual stimulus classification on GOD dataset}. The test set contains 50 categories that have no overlapping with the training set. The results for CADA-VAE, MVAE, MMVAE, MoPoE-VAE, and BraVL are taken from \protect\cite{du2023decoding}.}
     
    \resizebox{\linewidth}{!}{
    \centering
    \begin{tabular}{ccccccccccccccc}
        \toprule
        \multirow{2}*{Methods}& \multirow{2}*{Modality}&\multirow{2}*{Prompt}&\multicolumn{2}{c}{Subject 1} &\multicolumn{2}{c}{Subject 2} &\multicolumn{2}{c}{Subject 3} &\multicolumn{2}{c}{Subject 4} &\multicolumn{2}{c}{Subject 5}&\multicolumn{2}{c}{Average}\\
        \cmidrule(r){4-5}\cmidrule(r){6-7}\cmidrule(r){8-9}\cmidrule(r){10-11}\cmidrule(r){12-13}\cmidrule(r){14-15}
        &&&top-1 &top-5&top-1 &top-5&top-1 &top-5&top-1 &top-5&top-1 &top-5&top-1 &top-5\\
        \midrule
        CADA-VAE \cite{schonfeld2019generalized}&V\&T &- &6.31 & 35.70 &6.45 &40.12 &17.74 &54.34 &12.17&36.64&7.45&35.04&10.02&40.37\\ 
        MVAE \cite{wu2018multimodal} &V\&T &- &5.77 &31.51 &5.40 &38.46 &17.11 &52.46 &14.02&40.90&7.89&34.63&10.04&39.59\\   
        MMVAE \cite{shi2019variational} &V\&T &- & 6.63 &38.74 &6.60 &41.03 &22.11 &56.28&14.54&42.45&8.53&38.14&11.68&43.33\\
        MoPoE-VAE \cite{sutter2021generalized}&V\&T &- &8.54 &44.05 &8.34 &48.11 &22.68 &61.82 &14.57&58.51&10.45&46.40&12.92&51.78\\ 
        BraVL \cite{du2023decoding} &V\&T &- &9.11 &46.80  &8.91 &48.86 &24.00 &62.06 &15.08&60.00&12.86&47.94&13.99&53.13\\  
        \midrule
        BrainCLIP-Linear(ours)&V\&T & Text &10.00&34.00&22.00&56.00&\textbf{24.00}&60.00&16.00&58.00&14.00&58.00&17.20&53.20\\
         BrainCLIP-Linear(ours)&V\&T & CoOp &14.00&41.33&\textbf{24.00}&\textbf{62.67}&18.00&64.00&17.33&54.67&18.67&\textbf{58.67}&\textbf{18.40}&56.27\\
        BrainCLIP-VAE(ours) &V\&T &Text &8.00&42.00 &24.00&52.00&20.00&58.00&\textbf{20.00}&58.00&\textbf{20.00}&46.00&18.40&51.20\\
        BrainCLIP-VAE(ours) &V\&T &CoOp &\textbf{14.00}&\textbf{54.00} &20.00&60.00&21.33&\textbf{64.67}&16.67&\textbf{66.67}&18.00&52.00&18.00&\textbf{59.47}\\

      \bottomrule
      \end{tabular}}
       \label{zero_classfication}
\end{table*}

As we can see, BrainCLIP can work well with both two ways of prompting while CoOp outperforms manually designed prompts on the top-5 accuracy. And BrainCLIP integrated with CoOp outperforms the recently proposed models, such as
BraVL\cite{du2023decoding}, CADA-VAE\cite{schonfeld2019generalized}, MVAE\cite{wu2018multimodal}, MMVAE\cite{shi2019variational}, and MoPoE-VAE\cite{sutter2021generalized}, by a large margin. Note that their works are specifically designed for the zero-shot classification task while our work can be flexibly applied to different tasks. This is a huge advantage of our work. 

\begin{figure*}
  \centering
\centerline{\includegraphics[width=0.95\textwidth]{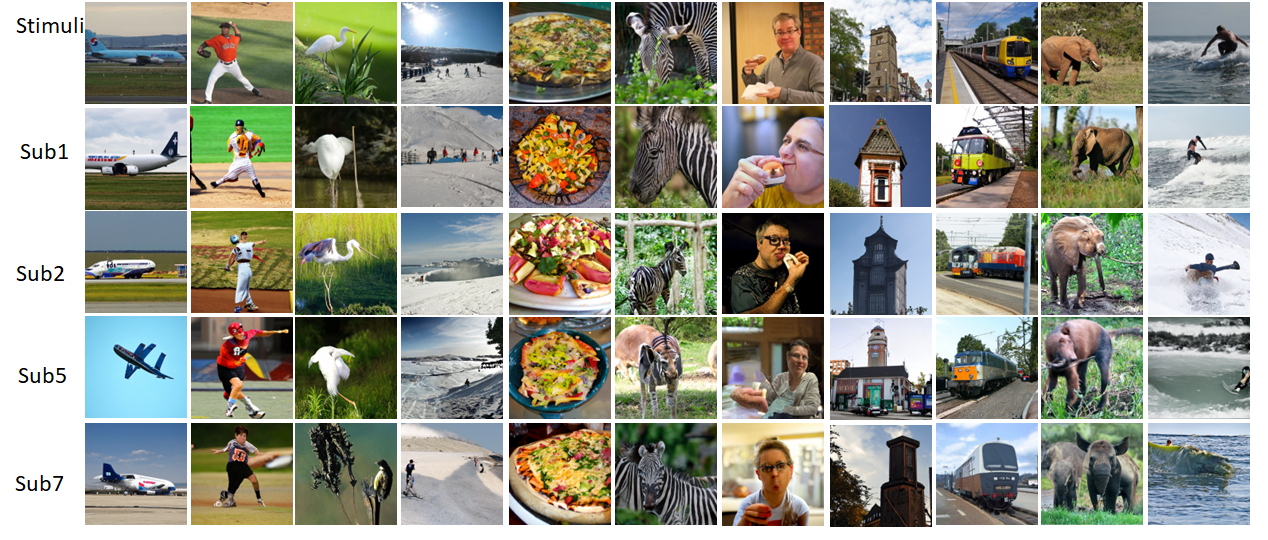}}
\caption{\textbf{Examples of fMRI Reconstructions from our BrainCLIP-VAE model}.  The first row shows the original visual stimuli. The following four rows show the reconstructions for each subject. }
\label{gen_all_sub}
\end{figure*}

\begin{table*}
    \setlength{\abovecaptionskip}{0pt}%
\setlength{\belowcaptionskip}{10pt}%
    \centering
      \caption{\textbf{Comparison with  state-of-the-art for visual stimulus reconstruction on NSD dataset}. Inception V3 is the 2-way identification based on the cosine similarity of their Inception V3 embeddings. CLIP is the 2-way identification of the image embeddings from the output layer of CLIP's visual encoder. Noised images are chosen by BrainCLIP from 170k images.}

    \begin{tabular}{cccccc}
        \toprule
        \multirow{2}*{Methods}& \multirow{2}*{Modality}&\multirow{2}*{Initial state}&\multicolumn{2}{c}{Similarity Measure}&\multirow{2}*{Average}\\
        \cmidrule(r){4-5}
        &&&Inception V3$\uparrow$& CLIP$\uparrow$\\
        \midrule
        Lin et al.\cite{lin2022mind} &-&-&78.2\%&-&-\\
        Takagi et al.\cite{takagi2022high}&-&- &76.0\%&77.0\%&76.5\%\\
        Brain-Diffuser\cite{ozcelik2023brain} &-&-&\textbf{87.2}\%&91.5\%&89.4\%\\
        \midrule
         BrainCLIP-Linear(ours)&V&Gaussian noise&73.3\%&83.6\%&78.5\%\\
        BrainCLIP-Linear(ours)&V\&T&Gaussian noise&79.7\%&89.4\%&84.6\% \\
         BrainCLIP-VAE(ours)&V&Gaussian noise&80.6\%&90.3\%&85.5\%\\
       BrainCLIP-VAE(ours)&V\&T&Gaussian noise&83.0\%&90.7\% &86.9\%\\
       BrainCLIP-Linear(ours)&V&noised image&82.9\%&93.3\%&88.1\%\\
       BrainCLIP-Linear(ours)&V\&T&noised image&85.0\%&94.6\% &89.8\%\\
        BrainCLIP-VAE(ours)&V&noised image&84.2\%&94.6\%&89.4\%\\
       BrainCLIP-VAE(ours)&V\&T&noised image&86.7\%&\textbf{94.8}\%&\textbf{90.8}\% \\
       
      \bottomrule
      \end{tabular}
       \label{gen_compare}
\end{table*}

\subsection{fMRI-to-image generation}
In this section, we evaluate BrainCLIP's fMRI-to-image generation ability on the NSD dataset(results on the GOD dataset are also reported, see Appendix). Some examples generated from fMRI under the guidance of our BrainCLIP-VAE models are displayed in Figure \ref{gen_all_sub}. The first row shows the original visual stimuli and the following four rows illustrate the reconstructions for each of the four subjects.
Since our work focuses on high-level semantic similarity, we do not expect the reconstructions to be identical to the visual stimuli at the pixel level. However, we find that the reconstructions have a high semantic fidelity to the original stimuli and show a robust consistency across subjects.  

 We also conducted a quantitative comparison with state-of-the-art to support the qualitative observations. In Table \ref{gen_compare}, we present the results of 2 image quality metrics. Inception V3 is the 2-way identification based on the cosine similarity of their Inception V3 embeddings(before the last FC layer,
the length-2048 vector). CLIP is the 2-way identification of the image embeddings from the output layer of CLIP. Both the two metrics focus on semantic similarity.  For
each generated image, we compare it with a randomly selected image and the ground truth one. Due to the randomness, we repeat the process 50 times.
Since BrainCLIP tends to  minimize the CLIP embedding distance directly, for a fair comparison, the CLIP embeddings used for training are extracted by CLIP RN101 while CLIP ViT-B/32 extracts those used for 2-way identification. 
All the results are averaged across 4 subjects.
We report the results for BrainCLIP under different settings to give a comprehensive evaluation. The following conclusions can be drawn from the results.
First, BrainCLIP has competitive performances under different settings in comparison with state-of-the-art; Second, on average, BrainCLIP-VAE performs better than BrainCLIP-Linear on fMRI-to-image generation task, which conforms with the trend shown in the results of fMRI-to-text retrieval task. Third, combining fMRI-image alignment and fMRI-text alignment can boost BrainCLIP's fMRI-to-image generation ability. Fourth, giving the diffusion model a meaningful initial state can improve reconstruction quality.

We need to point out that even though we have not emphasized pixel-level similarity in this work, it is possible to combine BrainCLIP with other pixel-wise stimulus reconstruction methods by using BrainCLIP as a global similarity constraint.


\section{Conclusion and future work}
  In this work, we proposed a task-agnostic fMRI-based brain decoding framework, BrainCLIP, which leverages the great cross-modal generalization ability of CLIP to connect brain activities with visual and textual data directly. Combined with the prompting techniques, BrainCLIP can achieve good zero-shot visual stimulus classification performance. Combined with the diffusion model, BrainCLIP establishes a new state-of-the-art for fMRI-based visual stimulus reconstruction in terms of semantic fidelity. 
  In the aspect of the training objective, our experiments indicate that the combination of visual and textual supervision can lead to better neural decoding accuracy for fMRI-text matching and fMRI-to-image generation. In the aspect of the model architecture, we explored a linear version and a VAE version for BrainCLIP's fMRI mapping network. Our Experiments demonstrate that BrainCLIP-VAE can obtain better fMRI-text retrieval and fMRI-to-image generation results on the NSD dataset, while BrainCLIP-Linear is better at fMRI-image retrieval.

  We should acknowledge that our current decoding accuracy is not perfect. Further improvements can start from the following aspects:
  1) to explore how to get better fMRI representation since the essence of brain decoding is to learn neural representation. 2) to combine our semantically-oriented decoding model with those pixel-wise reconstruction methods to simultaneously reconstruct low-level details and high-level semantic contents. These will be left for future work.


\section*{Acknowledgements}
This work was supported by National Natural Science Foundation of China (NO. 62088102), STI2030-Major Projects (NO. 2022ZD0208801), and China National Postdoctoral Program for Innovative Talents from China Postdoctoral Science Foundation (NO. BX2021239).

\bibliography{reference}
\clearpage
\appendix
\section*{Appendix}

\bibliographystyle{unsrt}
\section{Implementation details}
\setcounter{table}{0}   
\setcounter{figure}{0}
\renewcommand{\thetable}{A\arabic{table}}
\renewcommand{\thefigure}{A\arabic{figure}}

\subsection{Training of BrainCLIP-VAE}
\begin{figure}[h]
  \centering
  \includegraphics[width=1.\textwidth]{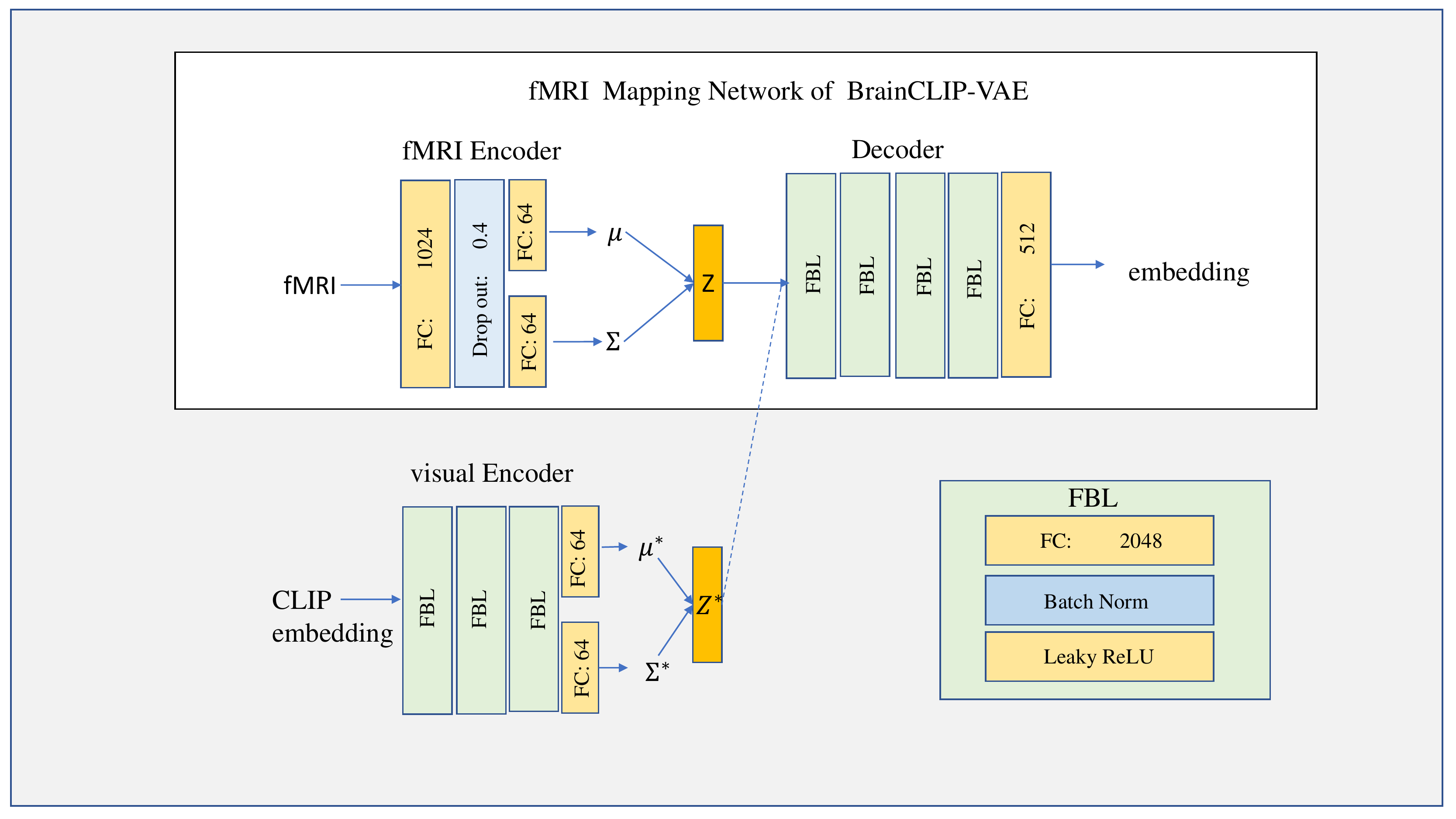}
  \caption{The architecture of our BrainCLIP-VAE model. The VAE decoder is jointly pre-trained with a visual encoder and then frozen. The integers shown in the blocks are the output dimensions.}
  \label{brainclip_vae}
\end{figure}

The architecture of the fMRI mapping network used in our BrainCLIP-VAE model is shown in Figure \ref{brainclip_vae}. Before training the fMRI mapping network, the VAE decoder was pre-trained with large-scale unpaired images from ImageNet and MSCOCO and then  frozen. During pre-training, the reconstruction loss  is the cosine distance; and the latent distribution is regularized by its Kullback-Leibler Divergence with a Gaussian Prior. The  Kullback-Leibler Divergence regularization weight used in this work is $0.001$ for both the pre-training process and the training of the fMRI Encoder. 

\subsection{Hyper-parameters}
The hyper-parameters were obtained through sweeps. The Mean recall of fMRI-to-image retrieval on the test set was used to select the trained models.  The batch size for the training of the fMRI encoder is 400 for the GOD dataset and 640 for the NSD dataset.  When combining visual and textual supervision, $\alpha$ is set to $0.5$. We updated the parameters using AdamW \cite{loshchilov2017decoupled} with $\beta1=0.9$, $\beta2 = 0.999$, $\epsilon=10^{-8}$, $lr=4.5\times10^{-5}$. 

On the GOD dataset, when training BrainCLIP-Linear, the weight decay is set to $2.0$, $\tau_1$ is set to $0.05$, and $\tau_2$ is set to $0.1$ for the 5 subjects; when training BrainCLIP-VAE,  the weight decay is set to $2.0$, $\tau_1$ is set to $0.05$, and $\tau_2$ is set to $[0.01,0,1,0.01,0.1,0.1]$ for the 5 subjects respectively.

On the NSD dataset, when training BrainCLIP-Linear, the weight decay is set to $3.0$, $\tau_1$ is set to $0.03$, and $\tau_2$ is set to $0.1$ for the 4 subjects; when training BrainCLIP-VAE,  the weight decay is set to $[35.0,25.0,35.0,25.0]$, $\tau_1$ is set to $0.01$, and $\tau_2$ is set to $[0.05,0.01,0.01,0.01]$ for the 4 subjects respectively. 

When conducting fMRI-to-image generation, the guidance scale $s$ is set to $1000$.
Training and inference are conducted on a single GeForce RTX 3090 GPU. Our source code will be publicly released later at \url{https://github.com/YulongBonjour/BrainCLIP}.
\subsection{Text prompts}
The text prompts used for zero-shot classification are listed as follows:

"a photo of a \{\}.",
        "a blurry photo of a \{\}.",
        "a black and white photo of a \{\}.",
        "a low contrast photo of a  \{\}.",
        "a high contrast photo of a  \{\}.",
        "a bad photo of a  \{\}.",
        "a good photo of a  \{\}.",
        "a photo of a small \{\}.",
        "a photo of a big  \{\}.",
        "a photo of the  \{\}.",
        "a blurry photo of the  \{\}.",
        "a black and white photo of the  \{\}.",
        "a low contrast photo of the  \{\}.",
        "a high contrast photo of the  \{\}.",
        "a bad photo of the  \{\}.",
        "a good photo of the  \{\}.",
        "a photo of the small  \{\}.",
        "a photo of the big \{\}."   
        
The “\{\}” is replaced by a specific class name. Their embeddings are averaged to get the classification weights.

\section{Additional results }
\setcounter{table}{0}   
\setcounter{figure}{0}
\renewcommand{\thetable}{B\arabic{table}}
\renewcommand{\thefigure}{B\arabic{figure}}
\subsection{ Additional results for the NSD dataset}
\begin{figure}
  \centering
  \includegraphics[width=1.\textwidth]{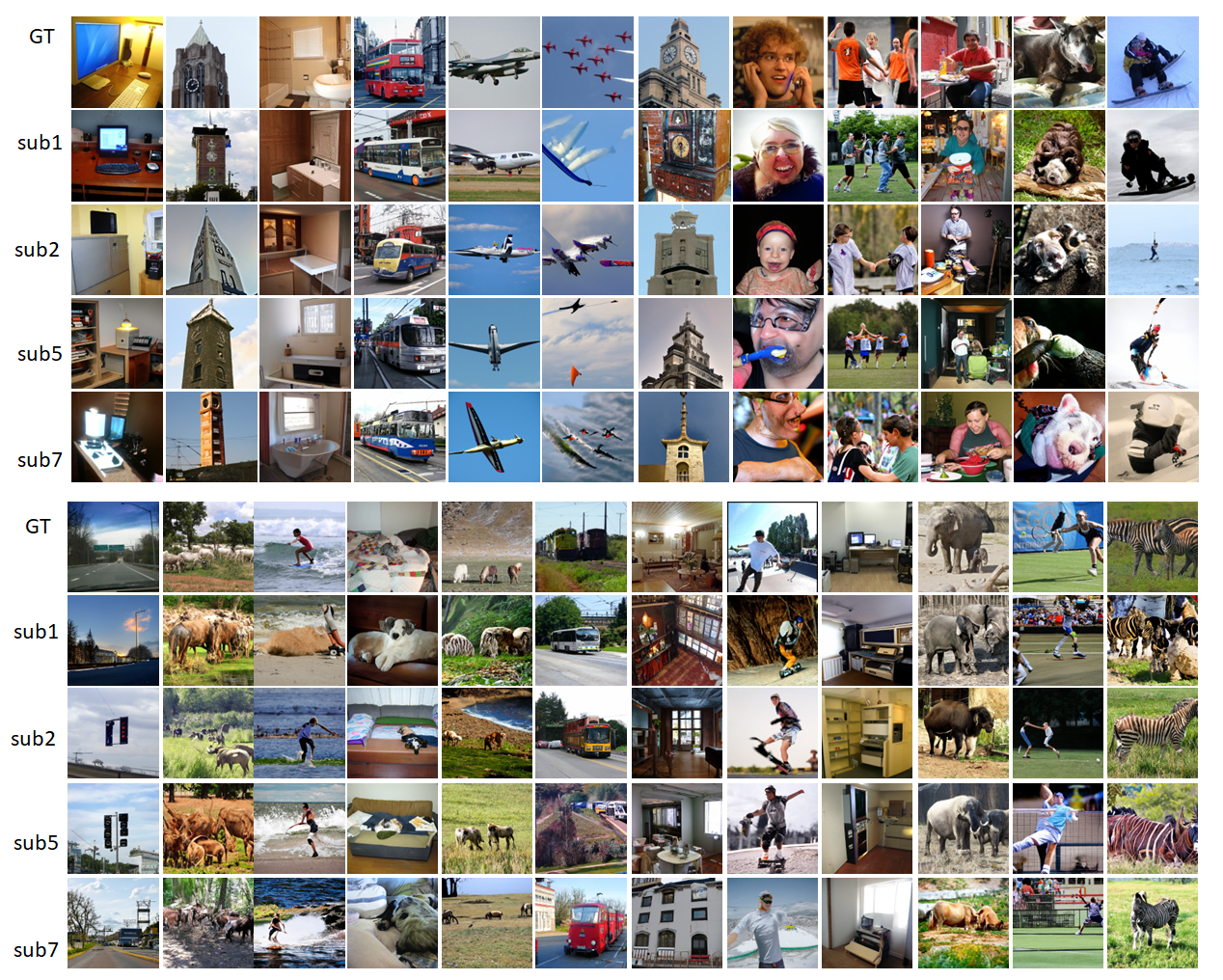}
  \caption{\textbf{Additional reconstructions for the NSD dataset. These samples are generated by BrainCLIP-VAE with a noised image as the initial state.}}
  \label{append_NSD_gen}
\end{figure}

In Figure \ref{append_NSD_gen}, we show more reconstructions for the NSD dataset.  
\subsection{fMRI-to-image generation on the GOD dataset}
\begin{figure}[h]
  \centering
  \includegraphics[width=0.6\textwidth]{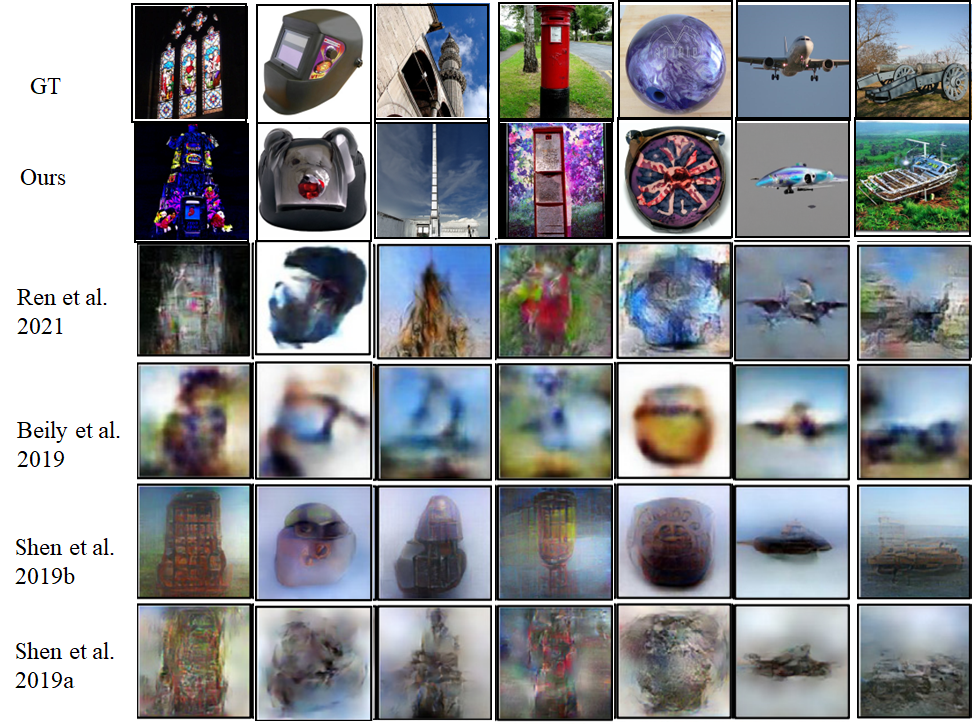}
  \caption{Comparison with state-of-the-art on GOD dataset  }
  \label{append_GOD_compare}
\end{figure}

\begin{figure}
  \centering
  \includegraphics[width=1.\textwidth]{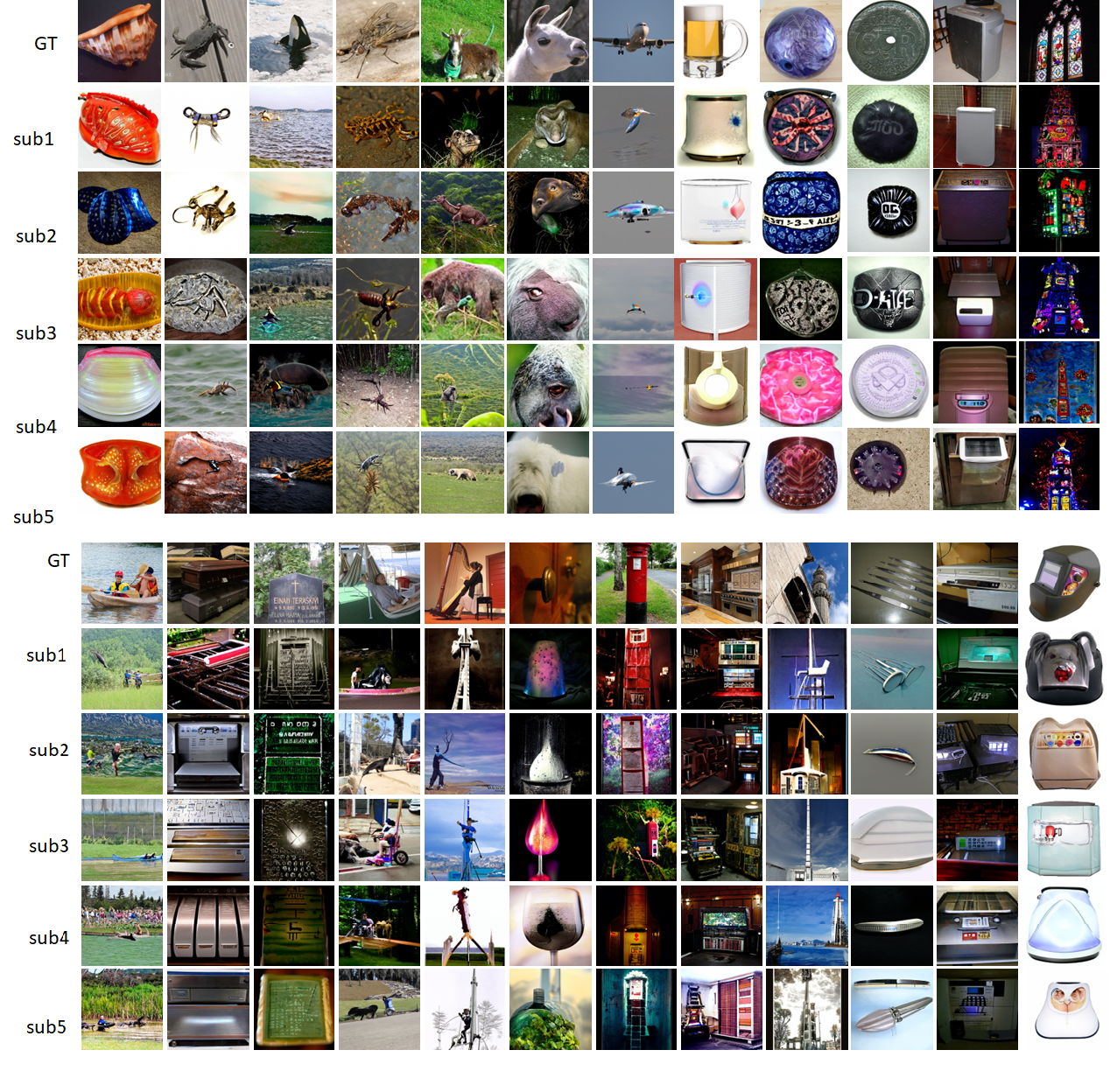}
  \caption{Reconstructions for the GOD dataset. These samples were generated by BrainCLIP-Linear with a random initial state.}
  \label{append_GOD_gen}
\end{figure}

BrainCLIP's quantitative performances for the stimulus reconstruction task on the GOD dataset are reported in Table \ref{append_gen_compare}. Since the test set of GOD is under a zero-shot setting, the categories are relatively harder to be decoded from fMRI patterns, we initialized the diffusion model with Gaussian noises instead of a selected image.  We found that combining visual and textual supervision can still improve decoding accuracy. However, BrainCLIP-VAE gets worse performances than BrainCLIP-Linear on the GOD dataset.  Since previous works on the GOD dataset haven't used the metrics used in this work, we only conduct a qualitative comparison with the state-of-the-art. In Figure \ref{append_GOD_compare}, we compare our reconstructions with those from Ren et al.\cite{ren2021reconstructing}, Beily et al.\cite{beliy2019voxels}, Shen et al.\cite{shen2019deep}, and Shen et al.\cite{shen2019end}. As we can see, our reconstructions are the most semantically recognizable. More reconstructions for the GOD dataset are shown in Figure \ref{append_GOD_gen}
\begin{table*}[h]
      \caption{\textbf{Quantitative results for stimulus reconstruction On the GOD dataset.}}
    \centering
    \begin{tabular}{cccccc}
        \toprule
        \multirow{2}*{Methods}& \multirow{2}*{Modality}&\multirow{2}*{Initial state}&\multicolumn{2}{c}{Similarity Measure}&\multirow{2}*{Average}\\
        \cmidrule(r){4-5}
        &&&Inception V3$\uparrow$& CLIP$\uparrow$\\
        \midrule
         BrainCLIP-Linear(ours)&V&Gaussian noise&62.4\%&83.7\%&73.1\%\\
        BrainCLIP-Linear(ours)&V\&T&Gaussian noise&\textbf{65.1}\%&\textbf{84.2}\%&\textbf{74.7}\% \\
         BrainCLIP-VAE(ours)&V&Gaussian noise&61.4\%&75.9\%&68.7\%\\
       BrainCLIP-VAE(ours)&V\&T&Gaussian noise&63.8\%&79.1\% &71.5\%\\
       
      \bottomrule
      \end{tabular}
       \label{append_gen_compare}
\end{table*}
\clearpage

\end{document}